\documentclass[article,nojss]{jss}

\usepackage{amsfonts}
\usepackage{optidef}
\usepackage{subfig}
\usepackage{hyperref}
\usepackage[frozencache,cachedir=.]{minted}
\newminted{python}{%
}

\newcommand{\reals}{\mathbb{R}}
\newcommand{\ie}{\textit{i}.\textit{e}., }
\newcommand{\norm}[1]{\left|\left|#1\right|\right|}

\author{Brian Liu \\MIT ORC \And
        Miaolan Xie\\Cornell ORIE \And Haoyue Yang \\Cornell Statistics \And Madeleine Udell \\Stanford MS\&E}
\title{\pkg{ControlBurn}:  Nonlinear Feature Selection with Sparse Tree Ensembles}

\Plainauthor{Brian Liu, Miaolan Xie, Haoyue Yang, Madeleine Udell} 
\Plaintitle{ControlBurn:  Nonlinear Feature Selection with Trees in Python.} 
\Shorttitle{Nonlinear Feature Selection with Trees} 

\Abstract{
\pkg{ControlBurn} is a Python package to construct feature-sparse tree ensembles that support nonlinear feature selection and interpretable machine learning. The algorithms in this package first build large tree ensembles that prioritize basis functions with few features and then select a feature-sparse subset of these basis functions using a weighted lasso optimization criterion. The package includes visualizations to analyze the features selected by the ensemble and their impact on predictions.
Hence ControlBurn offers the accuracy and flexibility of tree-ensemble models and the interpretability of sparse generalized additive models.

\pkg{ControlBurn} is scalable and flexible: for example, it can use warm-start continuation to compute the regularization path (prediction error for any number of selected features) for a dataset with tens of thousands of samples and hundreds of features in seconds. For larger datasets, the runtime scales linearly in the number of samples and features (up to a log factor), and the package support acceleration using sketching. Moreover, the \pkg{ControlBurn} framework accommodates feature costs, feature groupings, and $\ell_0$-based regularizers. The package is user-friendly and open-source: its documentation and source code appear on \href{https://pypi.org/project/ControlBurn/}{PyPI} and \url{https://github.com/udellgroup/controlburn/}.

}
\Plainkeywords{keywords, comma-separated, not capitalized} 

\Address{
  Brian Liu \\
  Operations Research Center \\
  Massachusetts Institute of Technology\\
  E-mail: \email{briliu@mit.edu}
  }

\begin{document}


\section{Introduction}

Feature selection is commonly used to improve model interpretability, parsimony, and generalization. In the linear setting, methods such as the lasso \citep{tibshirani1996regression}, group lasso \citep{friedman2010note}, and elastic net \citep{zou2005regularization} are frequently used to obtain sparse models. These techniques are valued for their ease of use and computational efficiency. 
To fit a sparse nonlinear model with the lasso, modelers often resort to ad-hoc feature engineering strategies like binning features \citep{wu2016revisiting} or adding pairwise feature interactions \citep{nelder1972generalized}, which can explode the dimension of the problem and still underperform compared to tree-based models like XGBoost \citep{chen2016xgboost} or Random Forest \citep{breiman2001random}.

Feature selection is more challenging in nonlinear models. Wrapper-based feature selection algorithms, such as recursive feature elimination (RFE), are computationally expensive as the model must be retrained to evaluate each subset of features \citep{darst2018using}. Feature importance metrics derived from nonlinear models, such as mean decrease in impurity (MDI) importance for tree ensembles, can be biased \citep{zhou2021unbiased} and can fail when some features are correlated \citep{liu2021controlburn}: a group of correlated features splits the MDI score between them, and so the score for an important group of features may be suppressed below the importance threshold for every individual feature in the group.

In this paper, we present \pkg{ControlBurn}, an efficient algorithm for feature selection in nonlinear models that works well even with many correlated features. \pkg{ControlBurn} first builds a large tree ensemble out of simple trees that isolate the effects of important single features and small subsets of features.
It then chooses a subset of the trees that jointly use a small number of features
by solving a group lasso problem. 
The algorithm is fast for large-scale data and yields an interpretable model that identifies the effects of important individual features and pairs of features.  
An implementation of \pkg{ControlBurn} is available as an open-source package in the \proglang{Python} programming language.

The paper is organized as follows. Section \ref{Methodology} presents the \pkg{ControlBurn} algorithm and section \ref{software} provides a tutorial of the \proglang{Python} implementation. Additional capabilities of \pkg{ControlBurn} are presented in section \ref{capabilities} and an advanced application of \pkg{ControlBurn} to emergency room triage appears in section \ref{application}.

\section{Nonlinear feature selection with trees} \label{Methodology}

\pkg{ControlBurn} first builds a tree ensemble (forest), say, by bagging or by gradient boosting.
Performance is sensitive to the quality and diversity of the tree ensemble
and \pkg{ControlBurn} works best when each tree uses only a few features. 
We discuss detailed strategies for building good ensembles in section~\ref{treebuild}.
\pkg{ControlBurn} then seeks a subset of the trees (subforest) that jointly use a small number of features and that predict the response well. 
It finds this subforest by solving a weighted lasso optimization problem. 
\pkg{ControlBurn} can find models with different sparsity levels by varying the regularization parameter in the lasso problem
and can choose the optimal sparsity level to minimize cross-validated error.
Given the selected features, \pkg{ControlBurn} can fit a final (``polished'') tree ensemble on the selected features to debias the model compared to the results after the lasso fit. 

We now describe each step in greater mathematical detail.
We begin by discussing methods for sparsifying a forest and 
revisit methods for building appropriate forests in section~\ref{treebuild}.


\subsection{General framework}\label{generalframework}
Given $n$ output-response pairs $\{(x^{(i)}, y^{(i)})\}_{i=1}^n$, 
suppose we have constructed $T$ trees: predictors that map each sample $x^{(i)}$
to a prediction $a^{(i)}$,
which can be continuous (for regression) or binary (for classification).
Each tree $t=1,\ldots,T$ is associated with a vector of predictions $a^{(t)} \in \mathbb{R}^{N}$ and a binary vector $g^{(t)} \in \{0,1\}^{P}$ that indicates which features are used as splits in tree $t$.
\pkg{ControlBurn} works best when the \emph{ensemble} (set of trees) is reasonably diverse, i.e. each tree is split on a different subset of features. We discuss methods for building trees in \S\ref{treebuild}. 

Our goal is to choose a sparse weight vector $w \in \mathbb{R}^T$ so that 
a weighted sum of the predictions of the trees $\hat y = \sum_{t=1}^T w_t a^{(t)}$ matches the response $y$ as well as possible.
We measure prediction error according to the loss function $\ell: \reals \times \reals \to \reals$, for example:
\begin{itemize}
    \item (for regression) squared error: $\ell(\hat y, y) = \|y - \hat y\|^2$,
    \item (for classification) logistic loss:  $\ell(\hat y, y) =\sum_{n=1}^N \log(1+\exp(-y_n \hat y_n))$,
    \item (for classification) hinge loss: $\ell(\hat y, y) =\sum_{n=1}^N (1 - y_n \hat y_n)_{+} $.
\end{itemize}

For convenience, denote $A = [a^{(1)}, \ldots, a^{(T)}] \in \mathbb{R}^{N\times T} $ 
and $G = [g^{(1)}, \ldots, g^{(T)}] \in \mathbb{R}^{N\times T}$.
With this notation, $Aw$ gives the predictions of the weighted tree ensemble and 
for each feature $p \in \{1,\ldots,P\}$, $(Gw)_p$ is nonzero if and only if feature $p$ is used by the ensemble.
Since $w \geq 0$ and $G$ is binary, $Gw = \sum_{t=1}^T u_t w_t$, where $u_t$ counts the number of features used by tree $t$, \ie the number of nonzeros in $g^{(t)}$.

\pkg{ControlBurn} chooses weights $w$ to minimize the average loss $\frac{1}{N}\sum_{n=1}^N \ell(Aw, y_n)$ of the prediction $Aw$ compared to the response $y$,
together with regularization that controls the number of features selected according to 
regularization parameter $\alpha$.
In the case of square loss, the optimal weight vector $w^\star$ solves the regularized maximum likelihood estimation problem

\begin{mini!}
{w}{\frac{1}{N}\left\|y-\mathrm{~A} w\right\|_{2}^{2}+\alpha u^T w \label{opt1obj}}{\label{optimizationproblem1}}{}
\addConstraint{w\geq 0. \label{opt1c1}}
\end{mini!}
We say feature $p$ is selected if $(Gw^\star)_p$ is non-zero.

\cite{friedman2003importance} also proposed using the lasso to select trees from an ensemble.
Our problem differs by weighting each tree $t$ by the number of features $u_t$ used by the tree, in order to reduce the number of \emph{features} used by the ensemble. 
This model will tend to choose trees that use only a few features while preserving predictive power.

Finally, \pkg{ControlBurn} optionally fits another ensemble model of choice, 
say random forest, on the subset of selected features.
This step, which we call \emph{polishing}, debiases the predictions compared to the lasso predictions \citep{meinshausen2007relaxed}.

\subsection{Building trees} \label{treebuild}

\begin{figure}[h]%
    \centering
\subfloat[\centering \textit{Good} ensemble]{{\includegraphics[width=0.485\textwidth]{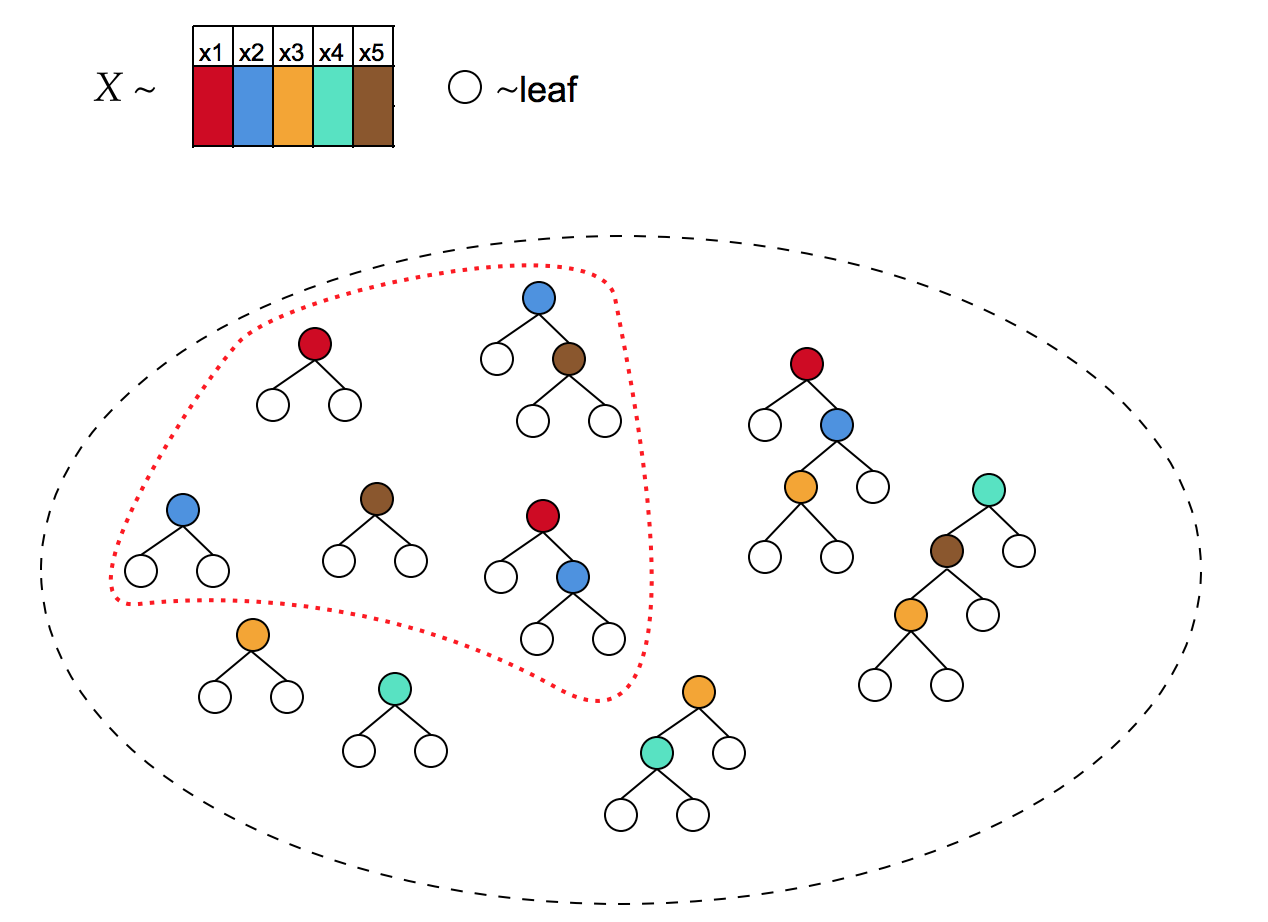} }}%
\qquad
\subfloat[\centering \textit{Bad} ensemble]{{\includegraphics[width=0.445\textwidth]{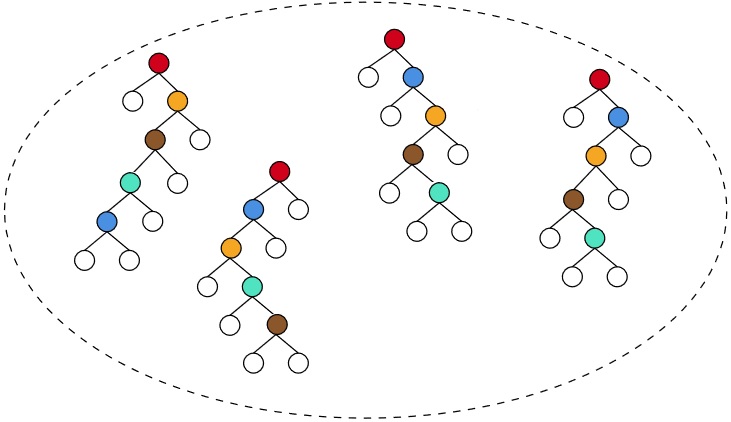} }}%
 \caption{\textit{Good} vs. \textit{Bad} ensembles for \pkg{ControlBurn}. The colors represent which features are used per split. In the \textit{good} ensemble, the red dashed line selects a feature sparse subforest; features $x_3$ and $x_4$ are excluded. In the \textit{bad} ensemble, selecting even a single tree selects all the features.}%
   \label{goodvbad}%
\end{figure}

The success of \pkg{ControlBurn} depends on the original tree ensemble $\{1 \ldots T \}$. The method works best when the original tree ensemble contains many trees that use only a small fraction of the total features so that the lasso problem can find feature-sparse subsets of the trees. For example, if $\{1 \ldots T \}$ is built via random forests \citep{breiman2001random} and each tree in the ensemble is grown to full depth, each tree $t \in \{1 \ldots T \}$ uses nearly every feature. As a result, \pkg{ControlBurn} will select either all or none of the features; there is no feature-sparse subforest (except the empty forest). Figure \ref{goodvbad} presents visualizations of \textit{good} vs. \textit{bad} ensembles to use in \pkg{ControlBurn}. \textit{Good} ensembles are diverse enough so that a feature-sparse subset of trees can be selected. In \textit{bad} ensembles, selecting a single tree selects almost all of the features.

Various algorithms to build ensembles with diverse trees are detailed in \cite{liu2021controlburn}. We summarize several such algorithms below.

\begin{itemize}
    \item \textbf{Incremental Depth Bagging:} Follow the bagging procedure proposed in \cite{breiman1996bagging} and grow trees of depth 1. When the training error of the ensemble converges, increment depth; repeat this procedure until the maximum depth is reached.
    
    \item \textbf{Incremental Depth Bag-Boosting:} Follow the incremental depth bagging procedure proposed above, but at each depth level, fit trees to the residuals of the model formed by the current ensemble. Across depth levels (boosting iterations) compute the out-of-bag (OOB) difference in error as a proxy for test error. Stop when the OOB error no longer improves.
    
    \item \textbf{Incremental Depth Double Bag-Boosting:} Follow the bag-boosting procedure detailed above, but when the training error of the ensemble converges, conduct a boosting iteration \emph{without} incrementing depth. When the OOB error between boosting iterations no longer improves, increase depth and repeat the procedure until the OOB error of the model no longer improves.
\end{itemize}

\subsection{Solving optimization problem 1}

Let $D \in \mathbb{R}^{T \times T}$ be a diagonal matrix such that the main diagonal of $D$ is equal to $u$. Each element of $u$, $u_t > 0$ represents the number of features tree $t$ uses. As a result, $D$ is positive-definite and invertible. We can rewrite Problem \ref{optimizationproblem1} as:

\begin{mini!}
{w}{\frac{1}{N}\left\|y-\mathrm{~AD^{-1}D} w\right\|_{2}^{2}+\alpha  \norm{Dw}_1 \label{opt2obj}}{\label{optimizationproblem2}}{}
\addConstraint{w\geq 0. \label{opt2c1}}
\end{mini!}

Let $x=Dw$; the above formulation is equivalent to:

\begin{mini!}
{w}{\frac{1}{N}\left\|y-\mathrm{~AD^{-1}} x\right\|_{2}^{2}+\alpha  \norm{x}_1}{\label{optimizationproblem3}}{}
\addConstraint{x \geq 0. \label{opt3c1}}
\end{mini!}

Problem 3 is equivalent to the non-negative garrote proposed in \cite{breiman1995better} and can be solved by existing lasso solvers found in \pkg{Scikit-learn} in \proglang{Python} or \pkg{glmnet} in \proglang{R}. The solution vector $x$ can be mapped back to the weights by a backsolve:
\[w = D^{-1}x. \]

Finally, the entire regularization path for $w$ can be computed efficiently by varying $\alpha$ and solving problem \ref{optimizationproblem3} with warm-start continuation \citep{friedman2010regularization}. This allows a practitioner to rapidly evaluate models with different feature sparsities. 

\section{Software} \label{software}

\subsection{Installation}

\pkg{ControlBurn} can be installed via the \proglang{Python} Package Index \href{https://pypi.org/project/ControlBurn/}{PyPI} and is available for Python 3.7 and above. The following dependencies are required.
\begin{itemize}
    \item \pkg{Numpy} \citep{harris2020array}
    \item \pkg{Pandas} \citep{jeff_reback_2022_6053272}
    \item \pkg{Scikit-learn} \citep{scikit-learn}
\end{itemize}

The source code for \pkg{ControlBurn} can be found in the following \href{https://github.com/udellgroup/controlburn}{repository}. To install \pkg{ControlBurn}, run the following command in terminal: \code{pip install ControlBurn}.

\subsection{Quick start}
Below, we present a quick example of \pkg{ControlBurn} on the classic Wisconsin breast cancer binary classification dataset \citep{wolberg1994machine}. Load \pkg{ControlBurn} in to your \proglang{Python} environment via the following.

\begin{pythoncode}
import ControlBurn
from ControlBurn.ControlBurnModel import ControlBurnClassifier
\end{pythoncode}

The code below initalizes a \code{ControlBurnClassifier}, grows a tree ensemble using the default method of incremental depth bag-boosting, \code{build_forest_method = 'bagboost'} and solves problem \ref{optimizationproblem1} with regularization penalty $\alpha = 0.1$.

\begin{pythoncode}
cb = ControlBurnClassifier(alpha = 0.1)
cb.fit(xTrain, yTrain)
features = cb.features_selected_

features
>>> ['mean concave points', 'worst perimeter', 'worst concave points']
\end{pythoncode}

During the ensemble building stage, 40 trees are grown and after the lasso step, 11 trees are selected. The selected ensemble is feature-sparse; the 11 trees only use the features \code{mean concave points, worst area,} and \code{worst concave points}. Only 3 of the 30 features in the full dataset are selected.

During the \code{fit} call, \pkg{ControlBurn} fits a polished model on the selected features. In this example, the default \code{polish_method = RandomForestClassifier()} is used. The predictions of the polished model on the selected features can be obtained by the following.

\begin{pythoncode}
predicted_classes = cb.predict(xTest)
predicted_probabilities = cb.predict_proba(xTest)
\end{pythoncode}

The prediction accuracy/AUC of the full 30 feature model is 0.96/0.99 and the prediction accuracy/AUC of the polished sparse model of 3 features is 0.96/0.99.  In two lines of code, we obtain a feature-sparse model that performs the same as the full model. \pkg{ControlBurn} closely follows \pkg{Scikit-learn} API conventions for easy integration into existing data science workflows. 

\subsection{Tutorial with interpretability tools} \label{cahousingtutorial}
In the following section, we use \pkg{ControlBurn} on the California Housing Prices regression dataset from UCI MLR \citep{Dua:2019}. Our goal is to select a sparse subset of features from \code{MedInc, HouseAge, AveRooms, AveBedrms, Population, AveOccup, Latitude,} and \code{Longitude} that jointly predict housing price well. We highlight the interpretability tools and features in the package.

To get started, run the following code.

\begin{pythoncode}
from ControlBurn.ControlBurnModel import ControlBurnRegressor
cb = ControlBurnRegressor(build_forest_method = 'doublebagboost', alpha = 0.02)
cb.fit(xTrain,yTrain)
prediction = cb.predict(xTest)
features = cb.features_selected_

features
>>> ['MedInc', 'HouseAge', 'Latitude', 'Longitude']
\end{pythoncode}

We fit a \code{ControlBurnRegressor} that uses incremental depth double bag-boosting to build the ensemble (and lasso to sparsify it). Double bag-boosting ensures that the effects due to single features are adequately represented before trees with two features are introduced, and similarly for each higher-order interaction.

Out of the 79 trees grown, only 16 are selected. This subforest only uses the features \code{MedInc, HouseAge, Latitude, Longitude}; only half of the features are selected.  The feature \code{MedInc} is the average earnings of households in the neighborhood surrounding a house, and \code{Latitude} and \code{Longitude} specify the location of the house. These features are important for predicting housing prices. The sparse polished model has a test mean-squared error (MSE) of 0.32 and the full model has a test MSE of 0.33. \pkg{ControlBurn} is able to quickly eliminate 4 redundant features in this example.

\pkg{ControlBurn} provides interpretability tools to analyze the selected subforest. Import the interpretability module and initialize an interpreter using the fitted \code{ControlBurnRegressor} object.

\begin{pythoncode}
from ControlBurn.ControlBurnInterpret import InterpretRegressor
interpreter = InterpretRegressor(cb,xTrain,yTrain)
\end{pythoncode}

To plot the feature importance scores of the selected subforest, run the following line of code.

\begin{pythoncode}
importance_score = interpreter.plot_feature_importances()
\end{pythoncode}

\begin{figure}[h!]
    \centering    
\includegraphics[width =  0.75\textwidth]{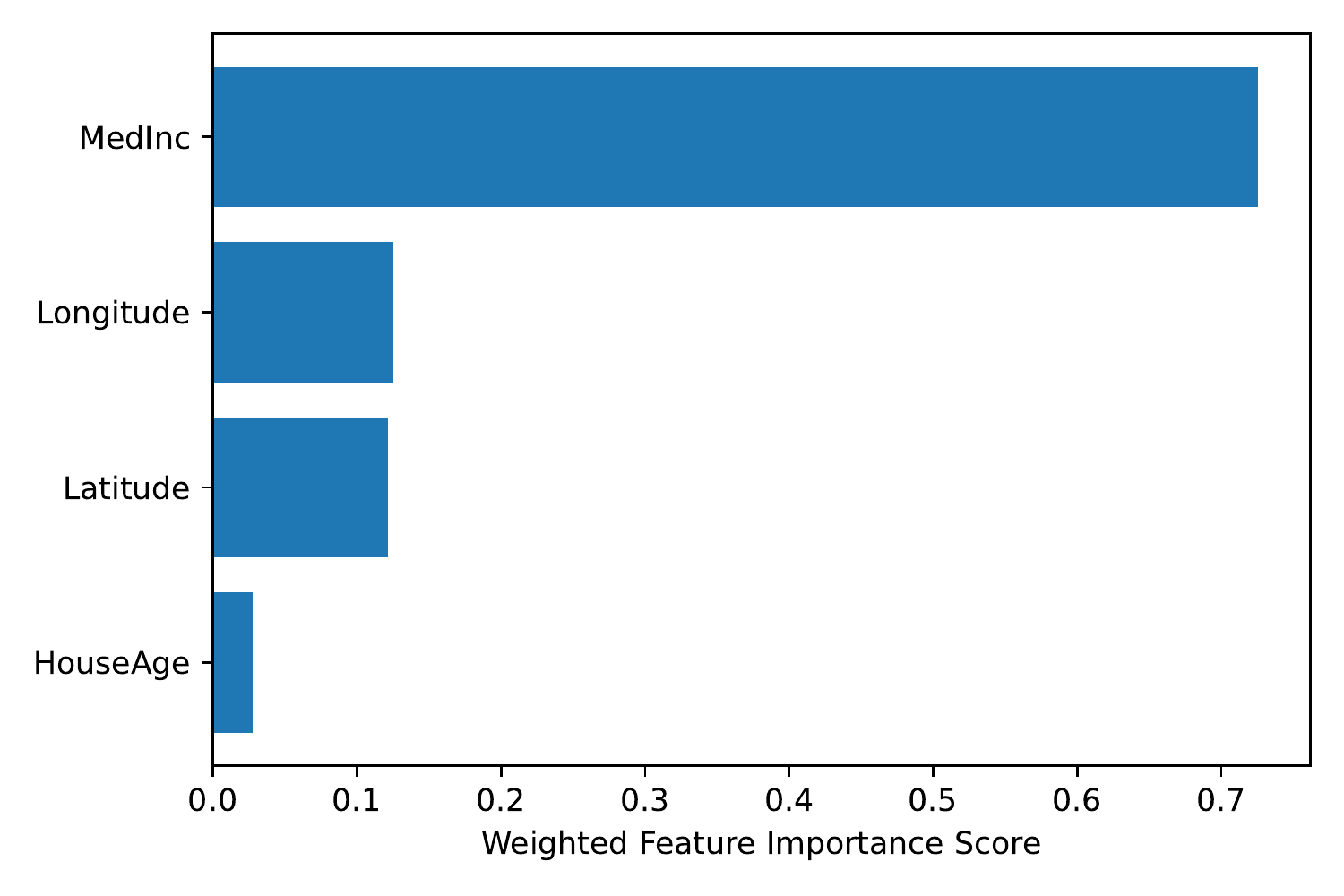}
    \caption{Weighted feature importance scores for the subforest selected by \pkg{ControlBurn}. }
    \label{featureimportances}
\end{figure}

The weighted feature importance scores in Figure \ref{featureimportances} are computed by multiplying the impurity-based feature importance score of each tree in the subforest by the weight the tree is assigned during the lasso step (Problem \ref{optimizationproblem1}).

The interpreter can also list the features used in each tree of the selected subforest.

\begin{pythoncode}
features_per_tree = interpreter.list_subforest(verbose = True)
\end{pythoncode}

This command outputs the following subforest structure:

\begin{pythoncode}
>>> ['MedInc'], ['MedInc'], ['MedInc'], ['MedInc'], ['MedInc'], ['MedInc'],
    ['MedInc'], ['MedInc'], ['MedInc'], ['MedInc'], ['MedInc'],  
    ['Latitude' 'Longitude'], ['Latitude' 'Longitude'], ['MedInc' 'HouseAge'],
    ['Latitude' 'Longitude'], ['Latitude' 'Longitude']
\end{pythoncode}

Each array shows the features used by a decision tree. In this example, the feature \code{MedInc} appears in several single-feature trees. We can use our interpreter to plot a shape function that shows how changes in the feature contribute to the response by aggregating these single feature trees.

\begin{pythoncode}
plot_single = interpreter.plot_single_feature_shape('MedInc')
\end{pythoncode}

\begin{figure}[h]
    \centering    
\includegraphics[width =  0.85\textwidth]{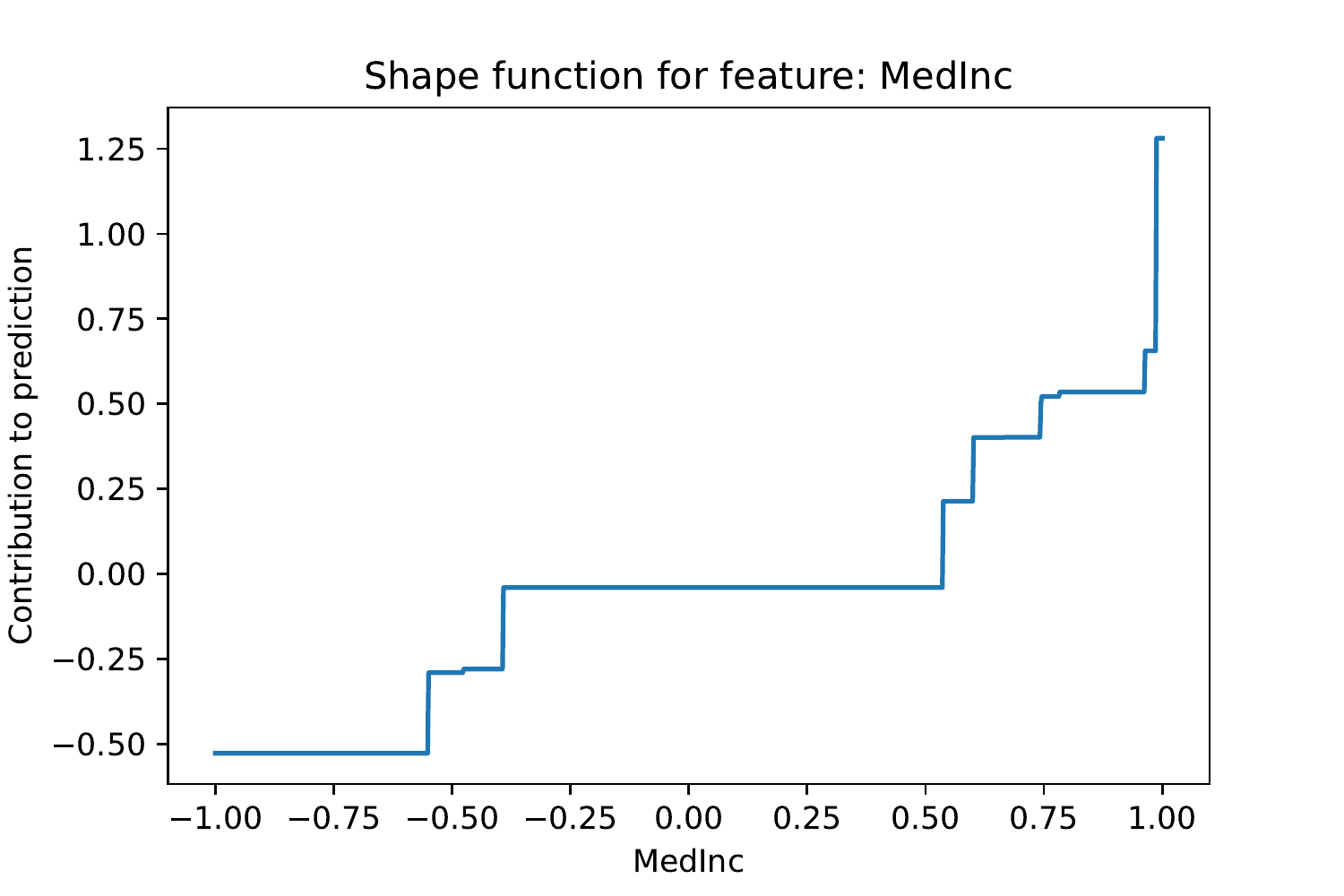}
    \caption{Shape function showing the contribution of feature \code{MedInc} towards the prediction.}
    \label{shapefunctionfigure}
\end{figure}

From the plot in Figure \ref{shapefunctionfigure}, we see that house prices rise with the median income of the neighborhood. We can also see the nonlinearity of the dependence: for example, the very steep rise in house prices as we move to the highest-income neighborhoods.

The features Latitude and Longitude also appear in many trees together. We can use the interpret module to examine how this pairwise feature interaction impacts the predictions. 
\begin{pythoncode}
plot_pairwise = interpreter.plot_pairwise_interactions('Latitude','Longitude')
\end{pythoncode}

\begin{figure}[p]
\centering
\includegraphics{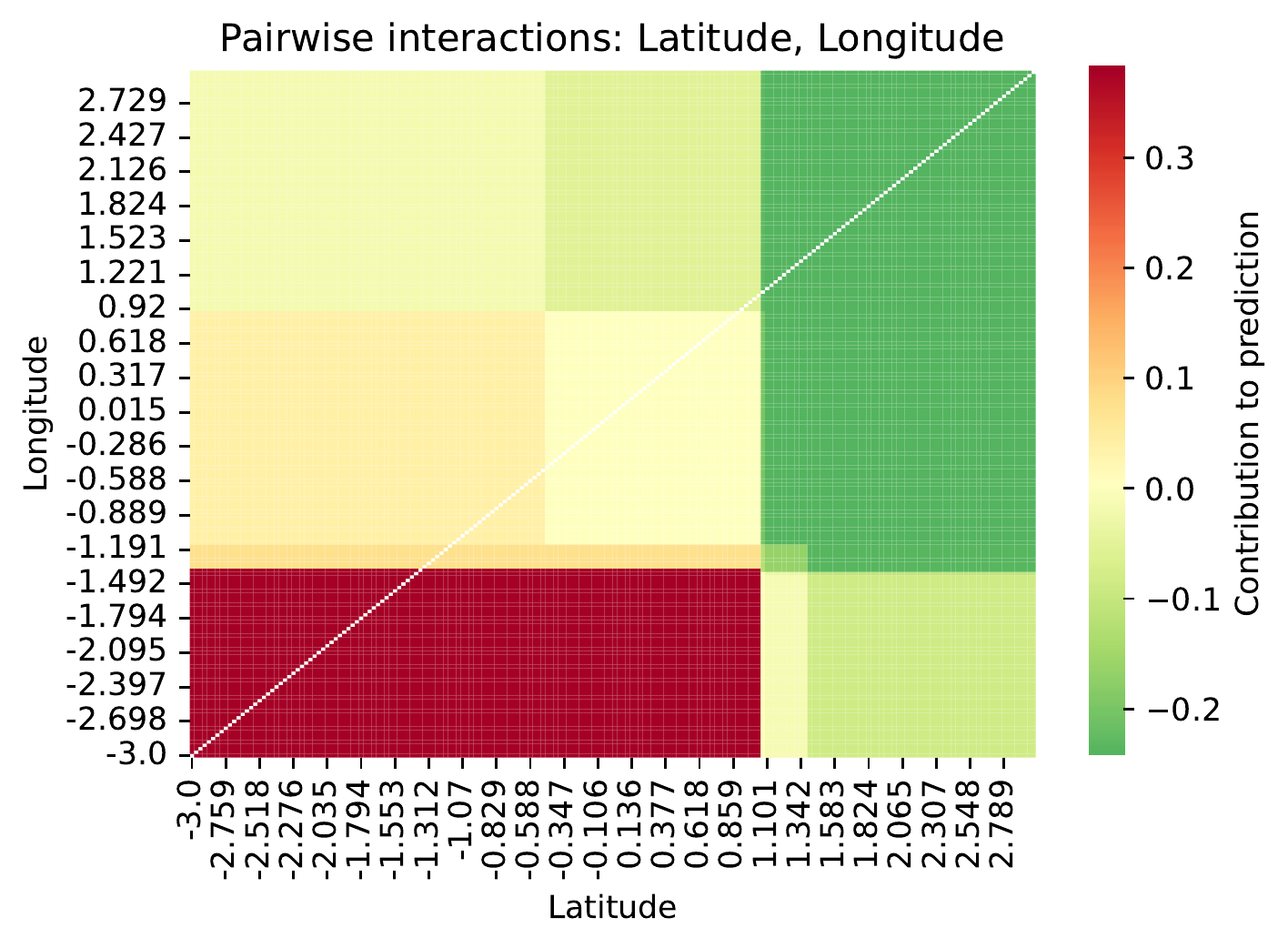}
\caption{Pairwise interaction between \code{Latitude} and \code{Longitude}.}
\label{pairwiseheatmap}
\includegraphics{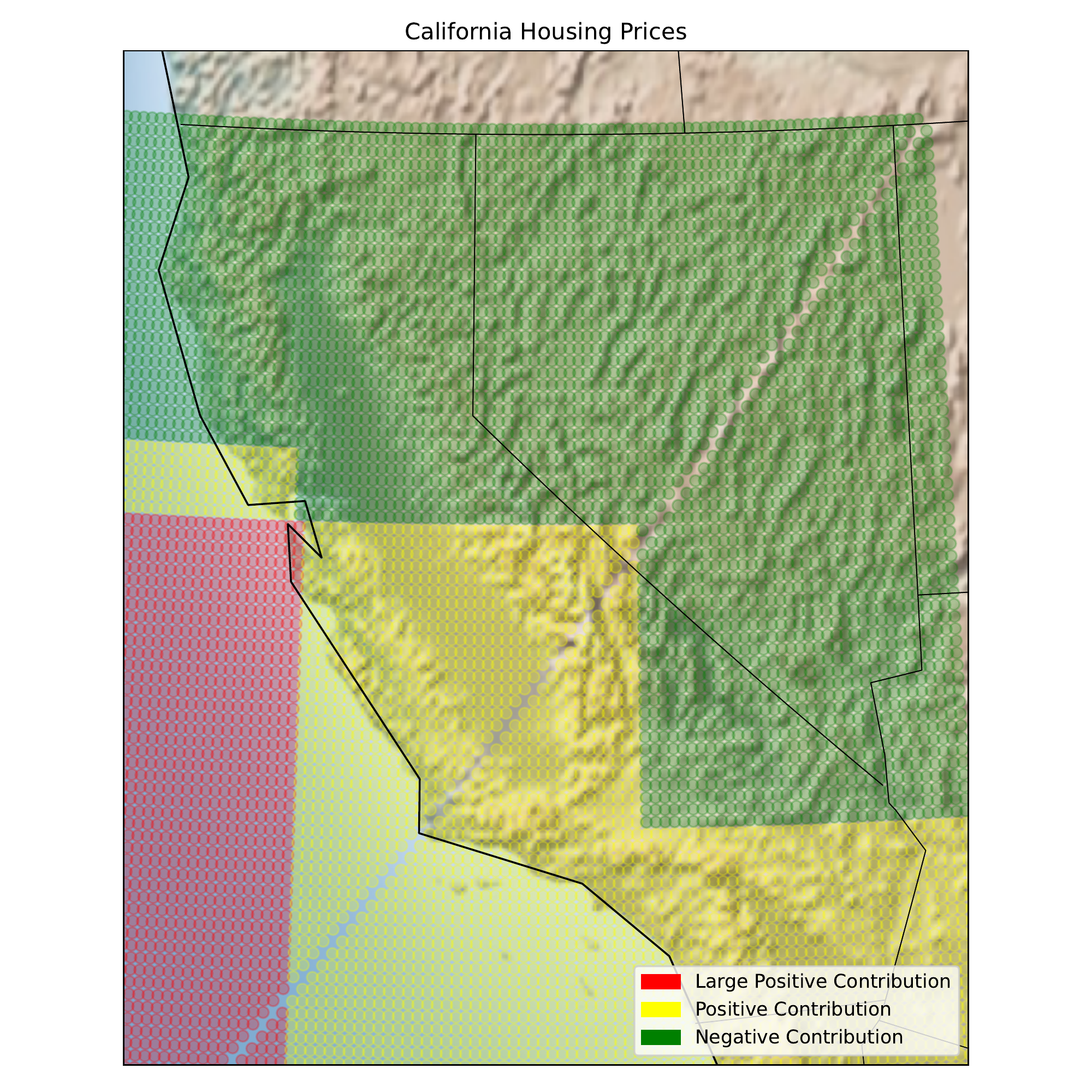}
\caption{\code{Latitude} and \code{Longitude} pairwise interaction effect on housing price, overlayed on a map of California.}
\label{CAmap}
\end{figure}

We observe in Figure \ref{pairwiseheatmap} that house prices are highest in the northwest of California, and lowest in the southeast. We can overlay this heatmap on a map of California to understand this effect better. In Figure \ref{CAmap} we observe that our model identifies that houses located in the San Francisco Bay area are most expensive; houses along the coast to Los Angeles and San Diego, in yellow, are next most expensive; and houses further inland, in green, are cheaper. These results are consistent with historical house price trends in the state.

\subsection{Regularization path}
By varying the regularization parameter $\alpha$, we can compute the entire regularization path and observe how features enter the support. The cost of computing the entire path is comparable to solving the lasso problem once. Execute the code below to compute the entire regularization path and plot how the identified feature importance of each feature changes as the regularization parameter alpha varies.

\begin{pythoncode}
alphas,coef = cb.solve_lasso_path(xTrain,yTrain)
regularization_importances = interpreter.plot_regularization_path()
\end{pythoncode}

\begin{figure}[h]
    \centering    
\includegraphics[width = \textwidth]{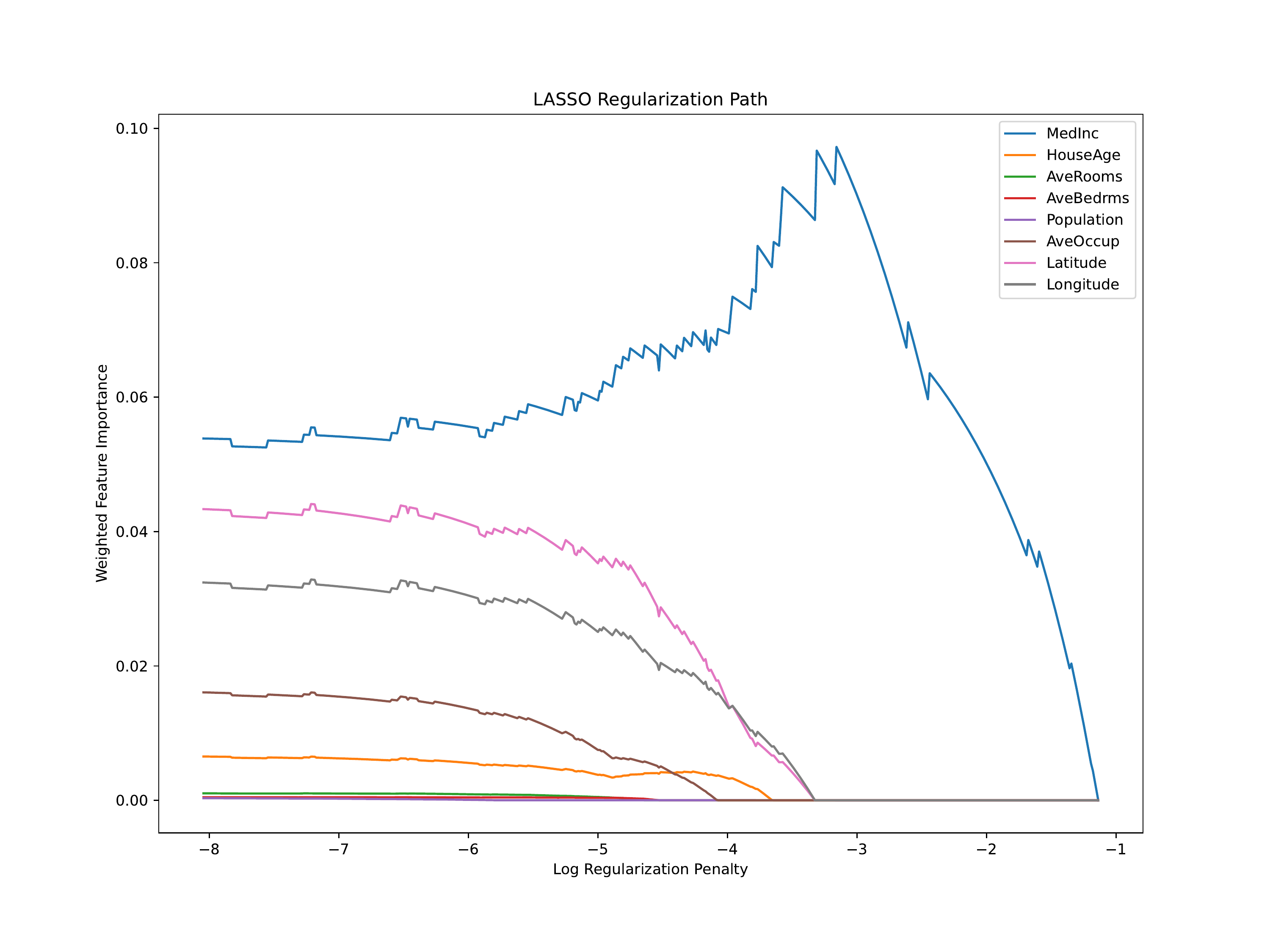}
    \caption{Regularization path obtained by varying $\alpha$. }
    \label{regularizationpath}
\end{figure}

The vertical axis of the plot in Figure \ref{regularizationpath} shows, for each feature, the MDI feature importance from each tree weighted by the lasso solution coefficient. Unlike the linear lasso coefficient regularization path, our feature importance paths are not necessarily monotonic. For example, in Figure \ref{regularizationpath} when \code{Latitude} and \code{Longitude} drop out of the subforest, the remaining feature \code{MedInc} is assigned a higher weight and therefore a higher weighted feature importance score.

\subsection{Selecting the best regularization parameter}

\pkg{ControlBurn} can automatically select a good regularization parameter by searching the regularization path for the parameter that minimizes the k-fold cross-validation error (default k = 5) of the model, using the \code{fit_cv} method. 
\begin{pythoncode}
best_alpha, support_size, best_features = cb.fit_cv(xTrain,yTrain, 
                                        show_plot = True, kwargs = {'tol':0.001})

best_alpha
>>> 0.012354087681630486
support_size
>>> 4
best feature sets
>>> ('AveOccup', 'Latitude', 'Longitude', 'MedInc'), 
    ('HouseAge', 'Latitude', 'Longitude', 'MedInc')
\end{pythoncode}

\begin{figure}[h]
    \centering    
\includegraphics[width = 1\textwidth]{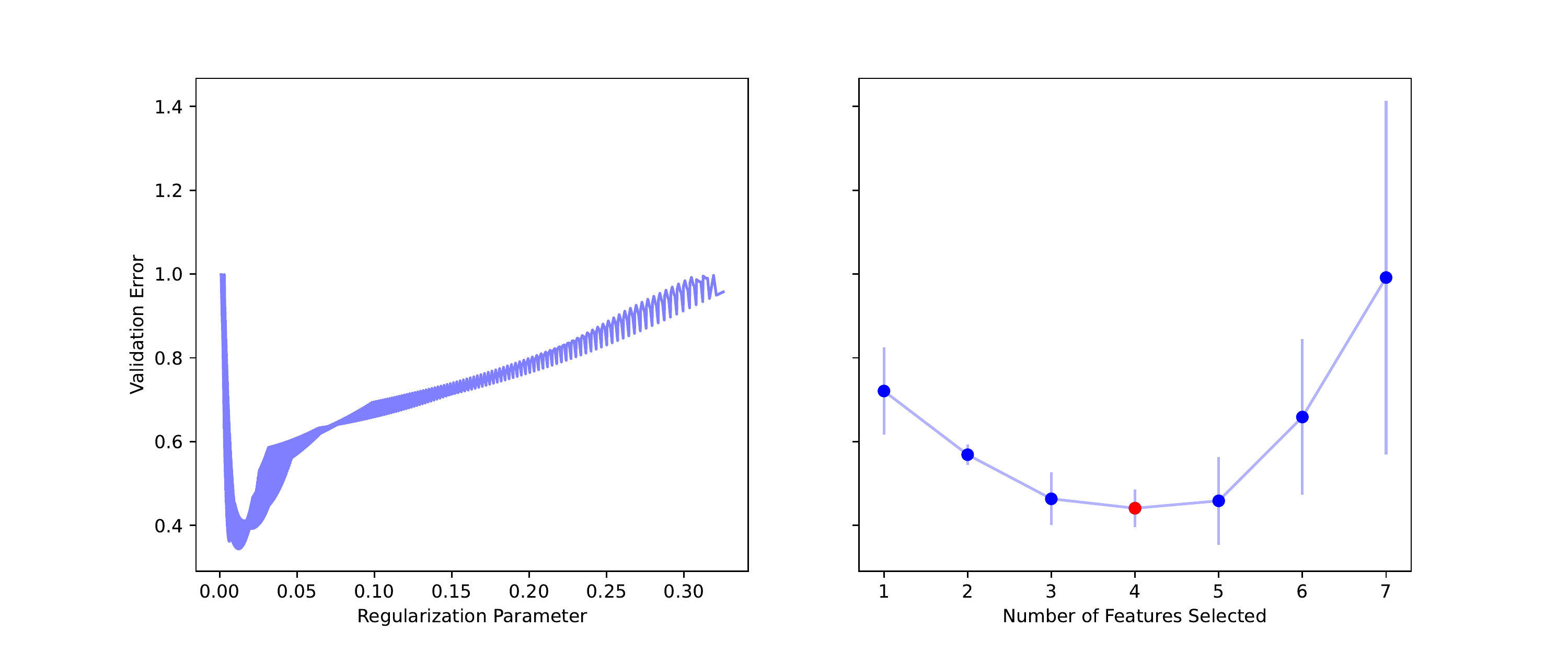}
    \caption{Left plot shows validation error vs. $\alpha$. Right plot shows validation error vs. number of features selected. The best support size contains four features.}
    \label{fitcvplot}
\end{figure}

In this example, \pkg{ControlBurn} selects a regularization parameter of 0.012, which selects four features. \code{Latitude}, \code{Longitude}, and \code{MedInc} are consistently selected as important features while \code{AveOccup} and \code{HouseAge} are variably included in the support in different folds. The \code{fit_cv} method automatically selects the optimal value of alpha and refits a tuned ControlBurnRegressor using that parameter. The features selected by the tuned model can be accessed with the command below.

\begin{pythoncode}
cb.features_selected_
>>> ['MedInc', 'HouseAge', 'Latitude', 'Longitude']
\end{pythoncode}

\section{Advanced capabilities}
\label{capabilities}

In the following section, we present some advanced capabilities of the \pkg{ControlBurn} package.

\subsection{Non-homogeneous feature costs}
In certain modeling applications, some features may be more expensive to obtain than others. 
In the \pkg{ControlBurn} framework, the user can assign each feature a cost and minimize the total cost of the selected features.

Let $c_p$ represent the relative cost of selecting feature $p$ and consider the framework presented in \S\ref{generalframework}. Let $\delta_t$ represent the set of features used by tree $t$. Assign each tree $t$ the following weight:

\[u_t= \sum_{p \in \delta_t} c_p.\]

The original \pkg{ControlBurn} framework with no feature costs corresponds to the case where $c_p = 1, \mkern9mu \forall \ p \in \{ 1 \ldots P \}.$

In the following example, we demonstrate how \pkg{ControlBurn} incorporates feature costs using the body fat regression dataset from \cite{penrose1985generalized}. The goal is to predict body fat percentage using the features: \code{Age,
 Weight,
 Height,
 Neck,
 Chest,
 Abdomen,
 Hip,
 Thigh,
 Knee,
 Ankle, Biceps, Forearm} and \code{Wrist}.
 
 Consider a hypothetical scenario where it is twenty times more time-consuming for an observer to measure a subject's torso, because the subject may need to remove bulky outerwear. We assign feature costs of 20 to \code{Chest, Abdomen, Hip} and feature costs of 1 to everything else. Assigning these feature costs and computing the entire regularization path with \pkg{ControlBurn} can be done via the following.
  
 \begin{pythoncode}
costs = [1,1,1,1,20,20,20,1,1,1,1,1,1]
cb = ControlBurnRegressor()
cb.solve_lasso_path(xTrain,yTrain,costs = costs)

\end{pythoncode}

Figure \ref{bodyfatfeaturecosts} compares the regularization paths computed by \pkg{ControlBurn} with and without feature costs. Without feature costs, \code{Abdomen} circumference is the best predictor of bodyfat percentage. When feature costs are included the trio, \code{Age, Height,} and \code{Neck} replace \code{Abdomen} as the most important features.

\subsection{Feature groupings}

In some settings, users choose whether or not to acquire costly \emph{groups} of features.
Once one feature in the group has been acquired, the rest are free.
Examples might include measurements taken as part of a complete blood panel (CBC): white blood cell counts are obtained at the same time as red blood cell counts.
As another example, in remote sensing, temperature, humidity, and pressure readings at a given location can be obtained by placing a single sensor. To create a model that is sensitive to these feature groupings, \pkg{ControlBurn} can penalize each tree for the \emph{groups}, not individual features, that it uses.

Consider the modeling framework presented in \S\ref{generalframework} and let $T_t$ represent the set of feature groups used by tree $t$. Let $c_g$ represent the cost of selecting group $g$ and assign each tree $t$ the weight
\[u_t= \sum_{g \in T_t} c_g.\]
The standard \pkg{ControlBurn} framework corresponds to the case where all groups are singletons and all weights $c_g$ are set to $1$.

In the section below, we return to the body fat regression example to demonstrate how \pkg{ControlBurn} can guide users' decisions on whether to acquire different feature groups. Consider the scenario where the features \code{Age, Weight, Height} can be obtained from the patient's medical history, and the remaining features are partitioned by their location on the patient's body. The features \code{Neck, Chest, Abdomen} can be obtained by measuring the patient's core, the features \code{Hip, Thigh, Knee, Ankle} can be obtained by measuring the patient's legs, and the features \code{Biceps, Forearm, Wrist} can be obtained by measuring the patient's arm. 

We can penalize selecting features over the four groups, History, Core, Legs, and Arms, each group having a cost of $1$, via the following.

 \begin{pythoncode}
groups = [1,1,1,2,2,2,2,3,3,3,4,4,4]
cb = ControlBurnRegressor()
cb.solve_lasso_path(xTrain,yTrain,groups = groups)
\end{pythoncode}

The list \code{groups} assigns each feature an integer group id, and in this setting \pkg{ControlBurn} minimizes the total number of groups selected.

\begin{figure}[h!] 
    \centering
    \includegraphics[width=\textwidth]{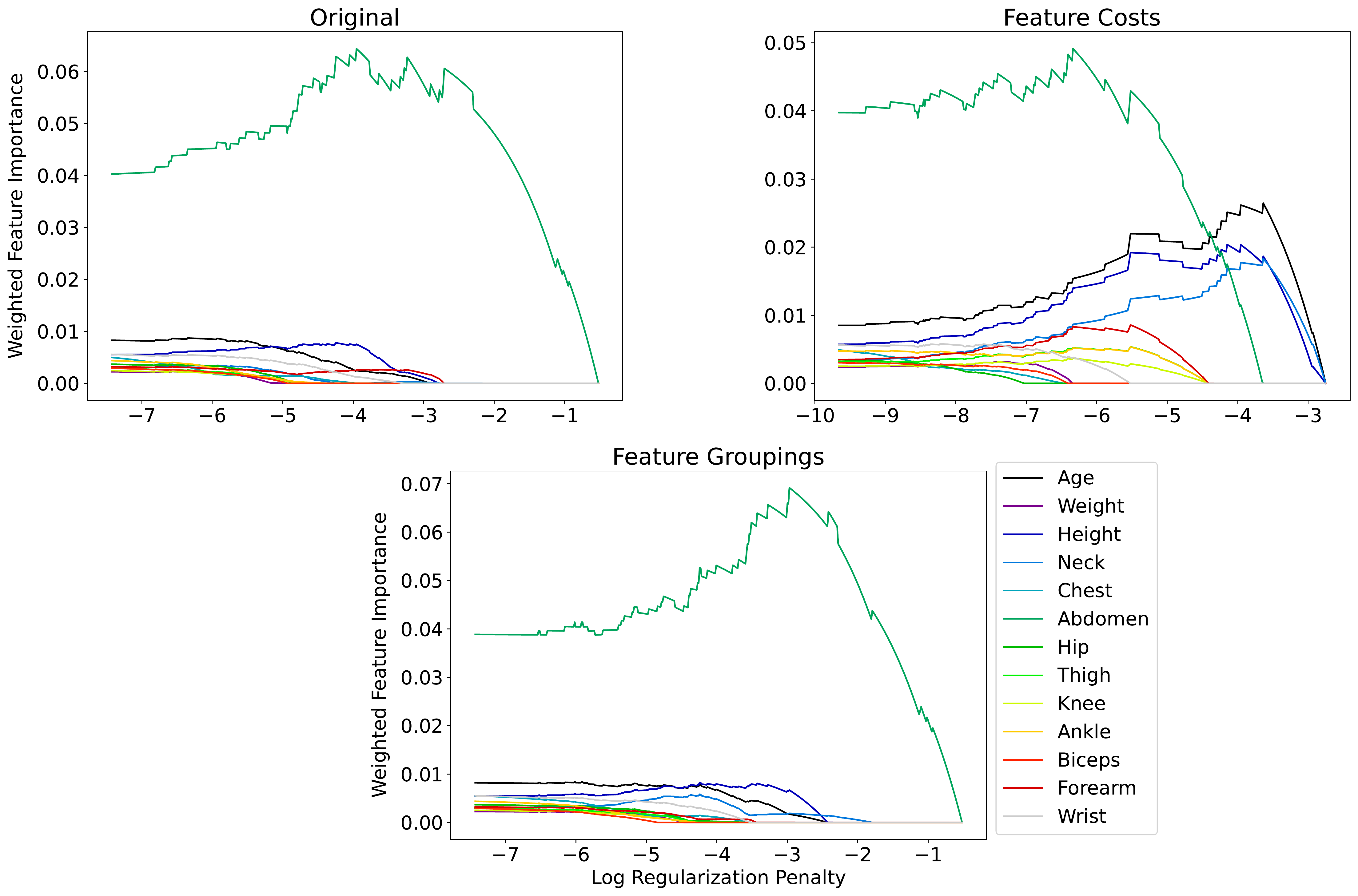}
    \caption{Effect of feature costs and feature groupings on \pkg{ControlBurn} regularization paths.}
    \label{bodyfatfeaturecosts}
\end{figure}

The bottom plot in Figure \ref{bodyfatfeaturecosts} shows the output of this code chunk. Note that compared to the original \pkg{ControlBurn} regularization path, the regularization path with feature grouping utilizes the feature \code{Neck} more frequently. This is due to the fact that the feature \code{Neck} is obtained at no additional cost when the feature \code{Abdomen} is selected since they both belong to the feature group Core. The feature group History is introduced shortly after the feature group Core.

To obtain the weighted feature importance scores of each group, run this command.

\begin{pythoncode}
group_names = ['History', 'Core', 'Legs', 'Arm']
group_importances = interpreter.plot_feature_importances(groups, 
                                group_names, show_plot = True)
\end{pythoncode}

\begin{figure}[h!] 
    \centering
    \includegraphics[width=.75\textwidth]{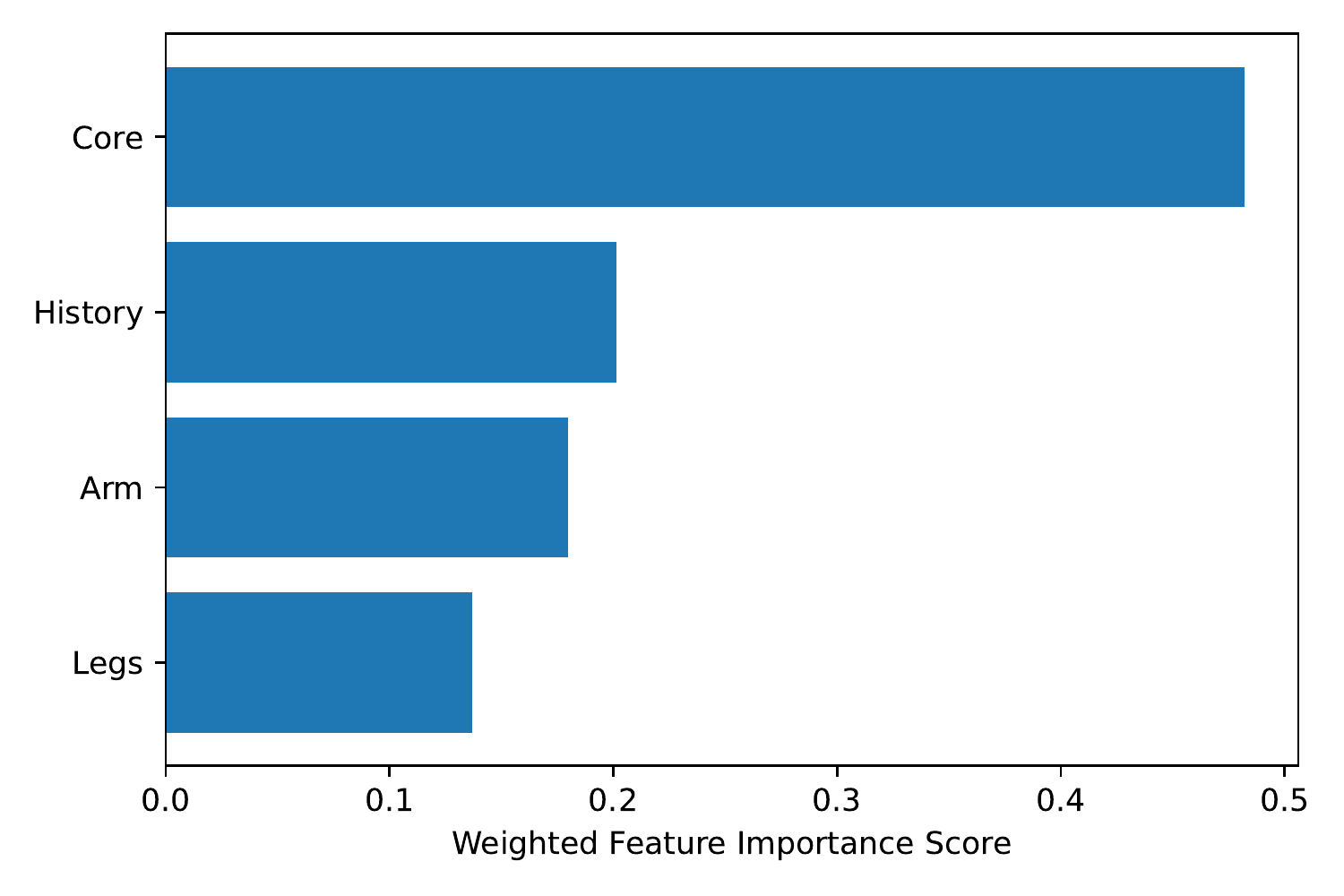}
    \caption{Plot of weighted feature importance scores by group.}
    \label{groupfeatureimportanceplot.fig}
\end{figure}

This outputs a bar plot of feature importance scores by group (Figure \ref{groupfeatureimportanceplot.fig}).

\subsection{Custom ensembles}

The ensemble building algorithms in \pkg{ControlBurn} can be easily replaced with custom pre-trained ensembles. Pre-trained ensembles are restricted to collections of \pkg{Scikit-learn} trees. For example, to train and parse in a \code{GradientBoostingRegressor}, run the following.

\begin{pythoncode}
from sklearn.ensemble import GradientBoostingRegressor
StochasticGB = GradientBoostingRegressor(max_depth = 2, max_features = 'log2', 
                                         subsample = 0.05)
StochasticGB.fit(xTrain.values,yTrain)
tree_list = np.ndarray.flatten(StochasticGB.estimators_)

cb_custom = ControlBurnRegressor(build_forest_method = 'custom', 
                                 custom_forest = tree_list)
cb_custom.fit(xTrain,yTrain)
\end{pythoncode}

Note that \pkg{ControlBurn} works best with diverse ensembles that use very few features per split and shallow trees. Caution should be taken that custom ensembles are adequately diverse. Otherwise \pkg{ControlBurn} may select only the null or the full model. 

\begin{figure}[h!] 
    \centering
    \includegraphics[width=\textwidth]{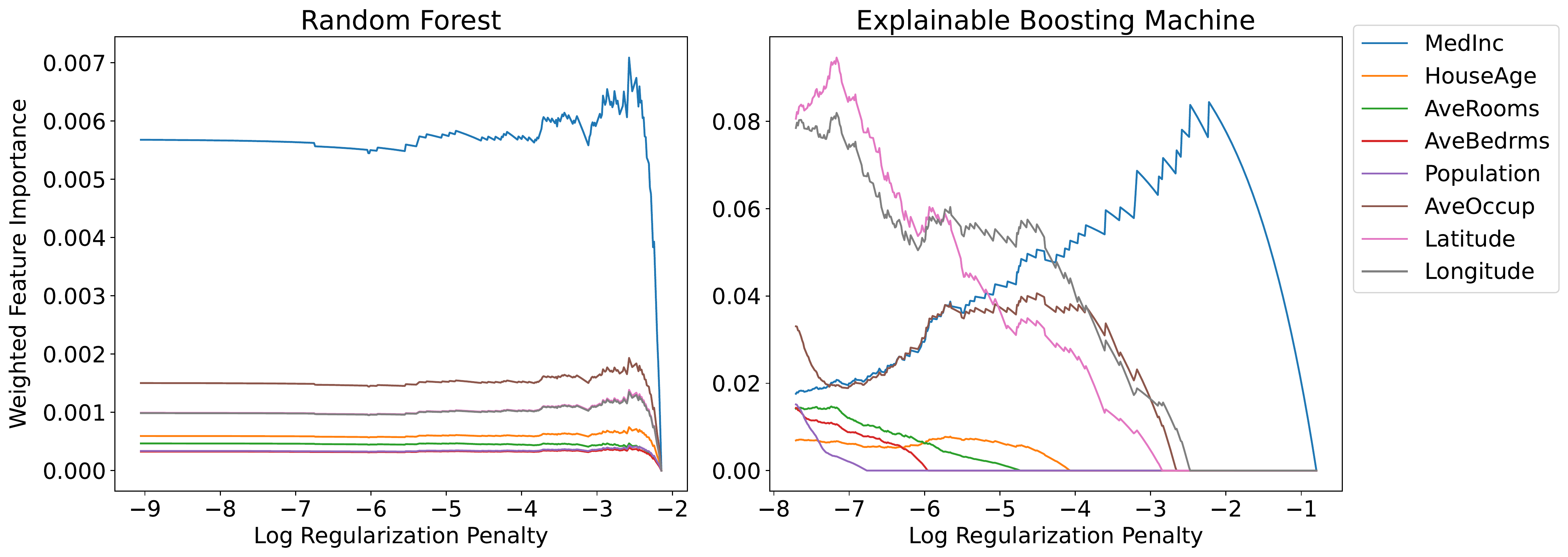}
    \caption{Regularization path of \pkg{ControlBurn} on custom ensembles. On a random forest, where each tree uses every feature, \pkg{ControlBurn} can select either the full or null model. The package works much better on EBMs and can select subforests of varying sparsities.}
    \label{rfvsebm}
\end{figure}

For example, given a Random Forest with deep trees, \pkg{ControlBurn} can only select a null or full model (see the left plot in Figure \ref{rfvsebm}); whereas the trees grown by an Explainable Boosting Machine (EBM) \citep{lou2012intelligible} each use at most two features, so \pkg{ControlBurn} works well.

\subsection{Sketching}

The optimization framework in \pkg{ControlBurn} scales linearly with the number of training observations, $N$. To reduce computation time, we can subsample/sketch matrix $A$. Define $S \in \mathbb{R}^{\eta \times N}$ as the Gaussian sketching matrix, where $\eta$ is the number of rows subsampled uniformly from $A$.  We rewrite problem \ref{optimizationproblem1} as

\begin{mini!}
{w}{\frac{1}{N}\left\|Sy-S\mathrm{~A} w\right\|_{2}^{2}+\alpha u^T w \label{sketchedobj}}{\label{sketchedproblem}}{}
\addConstraint{w\geq 0. \label{skectchedc1}}
\end{mini!}

To use sketching in the \pkg{ControlBurn} package, choose a proportion $\rho \in (0,1)$ of the training data to sample, which corresponds to $\eta = \lfloor N \rho \rfloor$ number of samples. For example, we may choose the sketching parameter $\rho = 0.1$ and run the following code:
\begin{pythoncode}
cb = ControlBurnRegressor()
cb.fit(xTrain,yTrain,sketching = 0.1)
\end{pythoncode}

\begin{figure}[h!] 
    \centering
 \includegraphics[width=.96\textwidth]{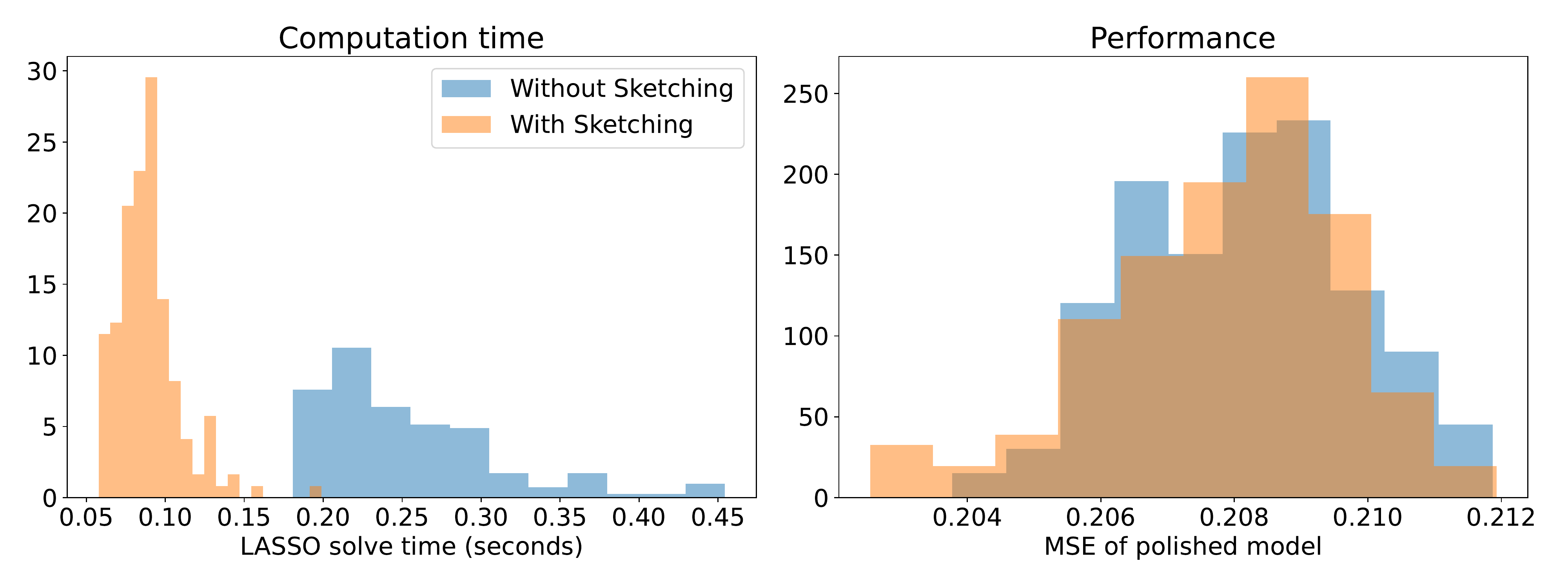} 
 \caption{Comparison of \pkg{ControlBurn} lasso solve with and without sketching. Sketching reduces computation time with negligible performance loss. }
 \label{sketchingcomparison}
\end{figure}

Figure \ref{sketchingcomparison} compares the runtime and performance of sketching versus no-sketching for \pkg{ControlBurn} on the California housing dataset. With a sketching parameter of $\rho = 0.1$, the optimization step of \pkg{ControlBurn} runs about 3x faster (left) and the selected model performs about equally well (right).

\subsection{Best subset selection}

\pkg{ControlBurn} can also use best-subset feature selection to select a feature-sparse subforest. In the linear setting, advancements in combinatorial optimization solvers have made best-subset selection feasible for medium-sized datasets (with thousands of samples and tens of features) \citep{bertsimas2016best}. 
On these datasets, best-subset selection has been shown to outperform lasso on regression problems 
in the high signal-to-noise ratio regime \citep{hastie2017extended,mazumder2020discussion}. 
One major advantage of best-subset selection is that the desired number of features can be directly specified. 
Let $K$ represent the desired number of features in the selected subforest. 
Best-subset selection over a tree ensemble finds weights $w_t$ for each tree $t \in \{ 1,\ldots,T \}$ to solve
\begin{mini!}
{w}{\frac{1}{N}\left\|y-\mathrm{~A} w\right\|_{2}^{2}\label{bestobj}}{\label{bestsubsetproblem}}{}
\addConstraint{\|Gw\|_0 = K \label{bestc1}}
\addConstraint{w\geq 0, \label{bestc2}}
\end{mini!}
where $\|\cdot\|_0$ counts the number of non-zero entries in its vector argument.
Constraint (\ref{bestc1}) ensures that exactly $K$ features are selected.
As in \S\ref{generalframework}, matrix $G$ captures which features are used by each tree. If an entry of $Gw \in \mathbb{R}^T$  is zero, the corresponding feature is excluded from the subforest. 

\pkg{ControlBurn} can choose features by best subset selection with the \code{solve_l0} function:
\begin{pythoncode}
cb = ControlBurnRegressor()
cb.bagboost_forest(xTrain,yTrain)
cb.solve_l0(xTrain, yTrain, K = 5)
cb.features_selected_
\end{pythoncode} 

\pkg{ControlBurn} uses \pkg{Gurobi} to efficiently solve the $\ell_0$-constrained mixed-integer quadratic program (MIQP) presented above \citep{gurobi}. 

\begin{figure}[h!] 
    \centering
 \includegraphics[width=.65\textwidth]{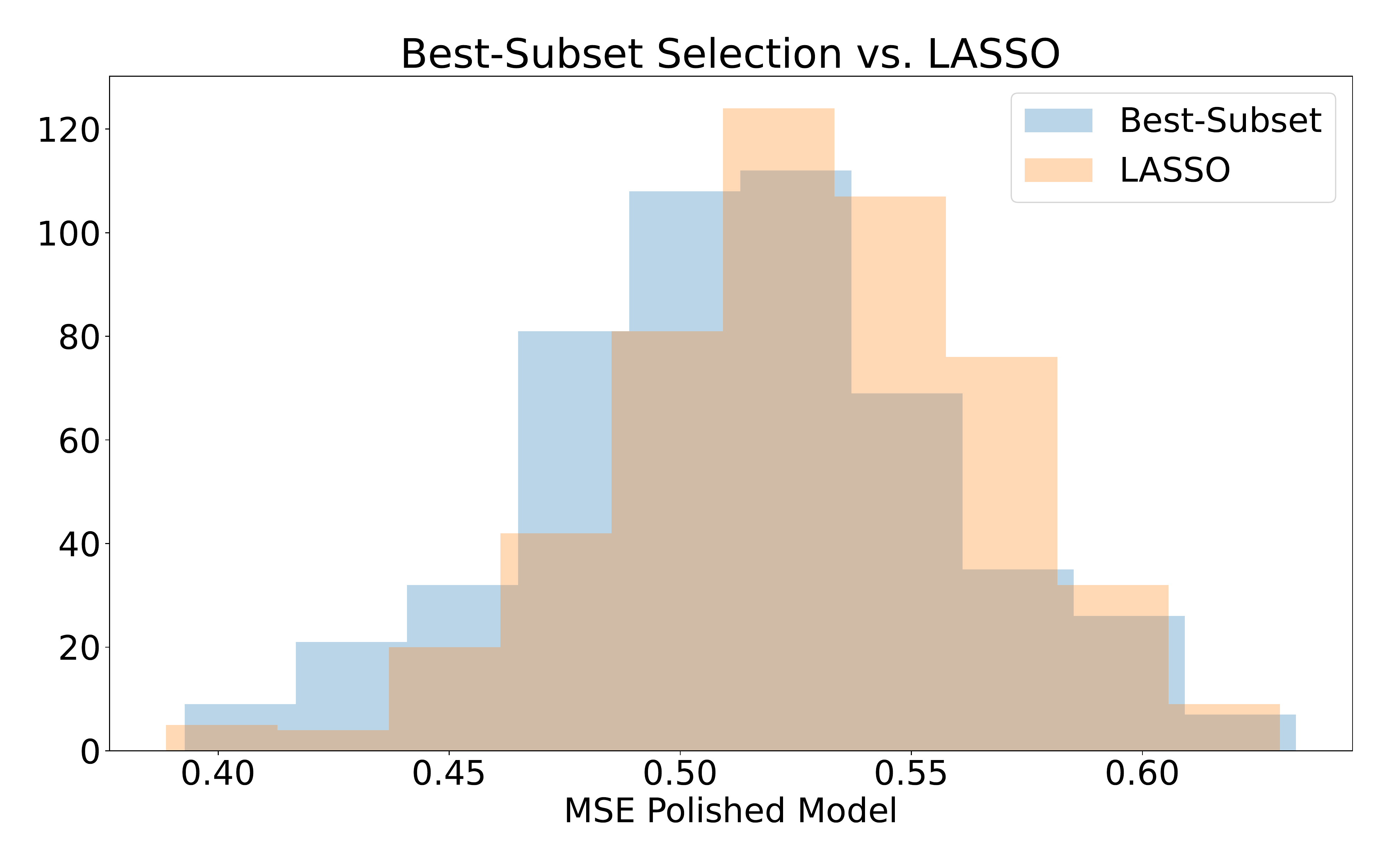} 
 \caption{\pkg{ControlBurn} with best-subset selection vs. lasso. Sparse models selected by best-subset selection improve test MSE slightly compared to those selected by lasso.}
 \label{bestsubsetcomparison}
\end{figure}

To demonstrate some advantages of \pkg{ControlBurn} with best-subset selection, we use best-subset selection and lasso to obtain the best 3-feature model on the US Crime dataset \citep{redmond2002data}. The goal is to predict neighborhood crime rates using 127 demographic and economic features, many of which are highly correlated. Figure \ref{bestsubsetcomparison} compares the distribution in performance between the best model selected by lasso versus the best model selected by best-subset selection. The best model selected by best-subset selection improves test MSE slightly compared to the best model selected via lasso, as is common in other problems \citep{mazumder2020discussion}.

\section{Real-world application: Emergency room triage}
\label{application}
We conclude by applying \pkg{ControlBurn} to a more extensive real-world example. The Yale Emergency Room Triage dataset \citep{hong2018predicting} consists of adult emergency room records from the Yale New Haven Health System from 2014 to 2017. In \cite{hong2018predicting}, the authors build binary classification models to predict hospital informations using triage information and patient medical history. Tree ensemble methods such as XGBoost achieve remarkable performance, with test AUC scores of 0.9. The feature importance scores of such models reveal that the emergency severity index (ESI) score, and what medications the patients were taking were the most influential predictors for hospital admission. ESI scores are triage scores assigned by the admitting medical staff and rank a patient on a scale from 1-5. Score 1 and 2 patients are in critical condition and require immediate life-saving care. Score 3 and 4 patients are non-critical but require care to stabilize. Score 5 patients require no emergency room resources to stabilize.

We use \pkg{ControlBurn} to select the most important predictors of hospital admission among non-critical patients that require care (ESI levels 3 and 4). The real-world implications of this classification task are interesting; patients with ESI scores 3 and 4 do not obviously need to be hospitalized but have the potential to take a turn for the worse. Determining which features predict hospital admission among this sub-population may provide useful clues to medical staff conducting triage.

During the ensemble-building stage, \pkg{ControlBurn} uses bag-boosting to build 112 trees that use 896 of the 969 features in the dataset. With a regularization parameter of $\alpha = 0.0015$, the lasso step of \pkg{ControlBurn} selects a 10 tree subforest that uses just 10 features. A random forest classifier fit on just the 10 selected features achieves a test AUC score of 0.81. The same classifier fit on the full feature space performs marginally better, with an AUC score of 0.88. For around a 10 percent decrease in performance, the number of features used in the model can be reduced by a factor of 90. Table \ref{tab:tablecompare} compares the performance of the model using features selected by \pkg{ControlBurn} against the full model, on various learning algorithms described in \citep{hong2018predicting}. Across all algorithms, the sparse model selected by \pkg{ControlBurn} performs within 10 percent of the full model.

\begin{table}[]
\centering
\begin{tabular}{|l|l|l|}
\hline
\textbf{Method}           & \textbf{Full Model AUC} & \textbf{Sparse Model AUC} \\ \hline
\textbf{Logistic Regression} & 0.88                    & 0.82                      \\ \hline
\textbf{Random Forest}       & 0.88                    & 0.81                      \\ \hline
\textbf{XGBoost}             & 0.90                    & 0.85                      \\ \hline
\end{tabular}
\caption{Comparison of AUC scores of full model vs. sparse model using features selected by \pkg{ControlBurn}. Methods from \cite{hong2018predicting} applied to our modified dataset: emergency room records filtered on ESI levels 3 and 4.}
\label{tab:tablecompare}
\end{table}

The subforest selected by \pkg{ControlBurn} has the following structure of 7 single feature trees and 3 multi-feature trees.

\begin{pythoncode}
>>> ['meds_gastrointestinal'], ['meds_cardiovascular'],['meds_cardiovascular'],
    ['meds_cardiovascular'], ['meds_cardiovascular'], ['meds_cardiovascular']
    ['meds_gastrointestinal']

    ['previousdispo' 'meds_cardiovascular' 'meds_gastrointestinal'],
    ['previousdispo' 'n_admissions' 'meds_cardiovascular' 'meds_gastrointestinal'],
    ['dep_name' 'age' 'insurance_status' 'n_admissions' 'triage_vital_sbp'
     'meds_analgesics' 'meds_cardiovascular' 'meds_vitamins']
\end{pythoncode}

The most important features, \code{meds_cardiovascular} and \code{meds_gastrointenstinal} are contained in single-feature, single-split trees. These numerical features indicate the amount of cardiovascular or gastrointestinal medication a patient is taking before they present at the ER.
From these single feature trees, it is apparent that patients currently taking these types of medications are more likely to be admitted; these medications are often used to treat chronic conditions in major organ systems.

\begin{figure}[h!] 
    \centering
 \includegraphics[width=\textwidth]{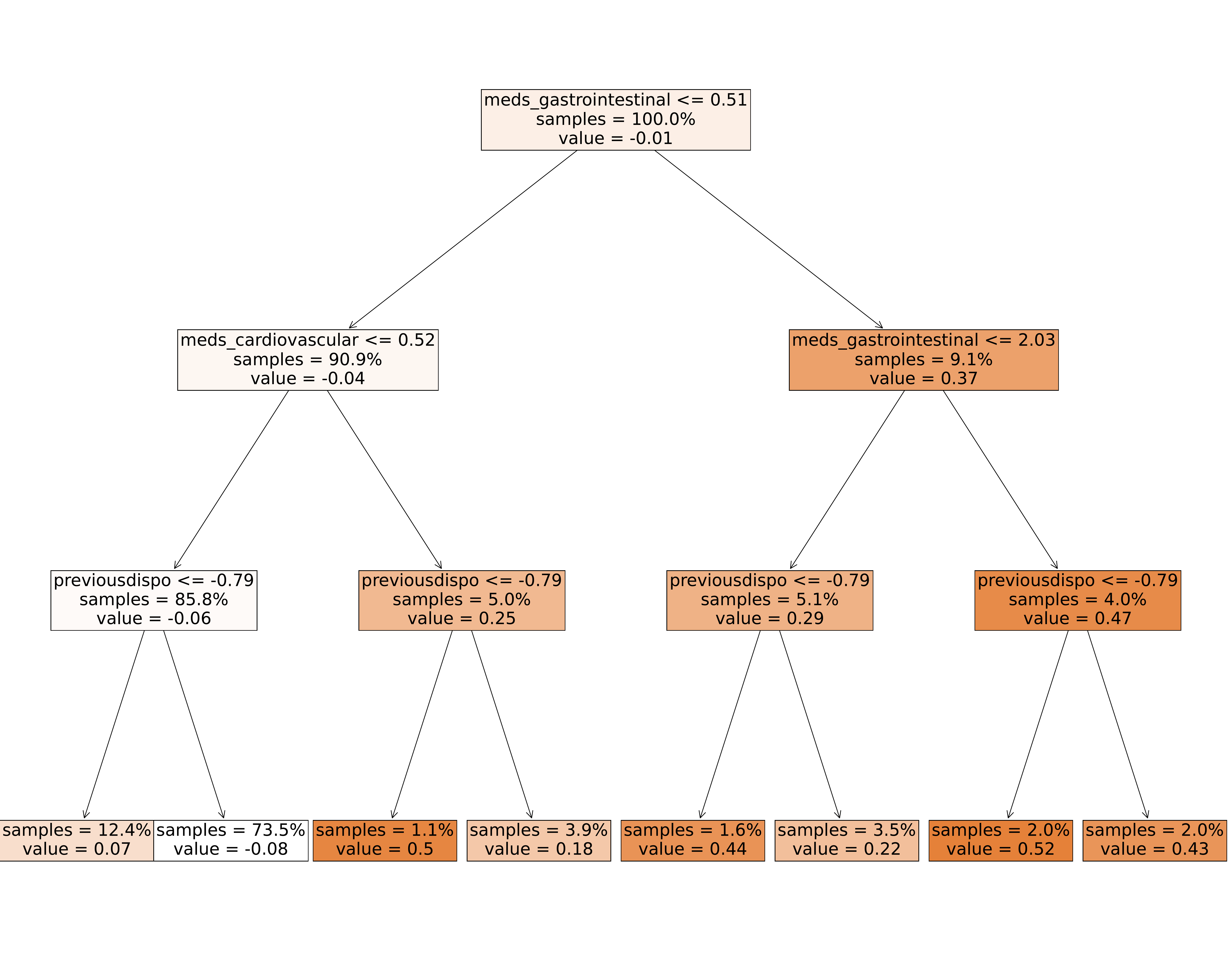} 
 \caption{Three feature tree selected by \textsc{ControlBurn} on the Yale emergency room triage dataset. Darker shades indicate the node increases the log-likelihood of hospital admission.}
 \label{yale3feat}
\end{figure}

In addition, the multi-feature trees selected reveal interesting interactions in the data. The 3-feature tree \code{[previousdisp, meds_cardiovascular, meds_gastrointenstinal]} is presented in Figure \ref{yale3feat}. The third layer of the tree splits on the ordinal feature \code{previousdispo} on the value -0.79. This feature represents the disposition of the patient on their last emergency room visit, patients with \code{previousdispo} values less than -0.79 were either admitted or left against medical advice. In both cases, the recommendation of the ER staff was to admit the patient and as a result, patients who have \code{previousdispo} values in these categories are more likely to be admitted upon subsequent ER visits. Consider the left-most branch in the tree in Figure \ref{yale3feat}. Patients along this branch present in a non-critical condition and are not currently taking cardiovascular or gastrointestinal medications, but are more likely to be admitted since they were admitted in the past. This can be due to an abundance of caution on the part of the ER staff.

 The features \code{n_admissions, dep_name, age, insurance_status, meds_vitamins,\\ meds_analgesics}, and \code{triage_vital_sbp} are introduced in the deeper trees selected. The structures of these trees are complex and more difficult to interpret but can reveal interesting relationships. For example, the feature \code{n_admissions} represents the number of prior hospital admissions for a patient and behaves similarly to the feature \code{previousdispo}; patients previously hospitalized are more likely to be admitted when visiting the ER. Patients with \code{insurance_status} equal to self-pay are less likely to be admitted since they may need to cover the cost of hospitalization. Finally, older patients are more likely to be admitted than younger ones.
 
By selecting a feature-sparse subforest, \pkg{ControlBurn} allows practitioners to identify important features and examine individual decision trees to determine how these features interact with each other and the response. The resulting subforest is much more interpretable than standard tree ensembles such as random forests, which contain hundreds to thousands of deep trees to visualize and may yield biased feature importance scores. In addition, a polished model fit on the features selected by the subforest often performs identically to an ensemble fit on the entire feature space. \pkg{ControlBurn} allows practitioners to extract insights from real-world data while preserving model performance.

\section{Concluding remarks}

The package \pkg{ControlBurn} extends linear feature selection algorithms to the nonlinear setting. The algorithm behind \pkg{ControlBurn} uses trees as basis functions and penalizes the number of features used per tree via a weighted $\ell_1$-penalty. By selecting a feature-sparse subforest, \pkg{ControlBurn} can quickly isolate a subset of important features for further analysis. \pkg{ControlBurn} also contains various built-in interpretability and visualization tools that can assist data analysis. By examining the structure and decision boundaries of the selected subforest, a practitioner can discover interesting insights in the data. In addition, \pkg{ControlBurn} can automatically evaluate the best support size by rapidly computing the entire path for the regularization parameter. Finally, \pkg{ControlBurn} is flexible and can accommodate various frameworks such as feature groupings and non-homogeneous feature costs. The package can also accept custom ensembles and an $\ell_0$-based solver for best-subset selection over trees. 
The source code for \pkg{ControlBurn} as well as the code and data to reproduce the experiments in this paper can be found at \url{https://github.com/udellgroup/controlburn/}. 

\bibliography{refs}

\end{document}